\documentclass{article}

\usepackage[preprint, nonatbib]{neurips_2024}

\usepackage[utf8]{inputenc} 
\usepackage[T1]{fontenc}    
\usepackage{hyperref}       
\usepackage{url}            
\usepackage{booktabs}       
\usepackage{amsfonts}       
\usepackage{nicefrac}       
\usepackage{microtype}      
\usepackage[table]{xcolor}         
\usepackage{graphicx}
\usepackage{multicol}
\usepackage{multirow}
\usepackage{makecell}
\usepackage{listings}
\usepackage{bm}
\usepackage{subfig}
\usepackage{amsmath}
\usepackage{amsfonts,amssymb}
\usepackage{wrapfig}
\usepackage{appendix}
\usepackage{listings}
\usepackage{xcolor}

\definecolor{keyword}{rgb}{0.0, 0.0, 1.0}
\definecolor{identifier}{rgb}{0.0, 0.5, 0.0}
\definecolor{comment}{rgb}{0.5, 0.5, 0.5}
\definecolor{string}{rgb}{0.6, 0.1, 0.1}

\title{Vision Transformer with Sparse Scan Prior}

\author{%
  Yuguang Zhang$^{1}$, Qihang Fan$^{1, 2}$, Huaibo Huang$^{1}$\thanks{Huaibo Huang is the corresponding author.}\\
  $^1$MAIS \& CRIPAC, Institute of Automation, Chinese Academy of Sciences, Beijing, China\\
   $^2$School of Artificial Intelligence, University of Chinese Academy of Sciences, Beijing, China\\
  \texttt{huaibo.huang@cripac.ia.ac.cn,}\\
}

\begin{document}

\maketitle

\begin{abstract}
  In recent years, Transformers have achieved remarkable progress in computer vision tasks. However, their global modeling often comes with substantial computational overhead, in stark contrast to the human eye's efficient information processing. Inspired by the human eye's sparse scanning mechanism, we propose a \textbf{S}parse \textbf{S}can \textbf{S}elf-\textbf{A}ttention mechanism ($\rm{S}^3\rm{A}$). This mechanism predefines a series of Anchors of Interest for each token and employs local attention to efficiently model the spatial information around these anchors, avoiding redundant global modeling and excessive focus on local information. This approach mirrors the human eye's functionality and significantly reduces the computational load of vision models. Building on $\rm{S}^3\rm{A}$, we introduce the \textbf{S}parse \textbf{S}can \textbf{Vi}sion \textbf{T}ransformer (SSViT). Extensive experiments demonstrate the outstanding performance of SSViT across a variety of tasks.  Specifically, on ImageNet classification, without additional supervision or training data, SSViT achieves top-1 accuracies of \textbf{84.4\%/85.7\%} with \textbf{4.4G/18.2G} FLOPs. SSViT also excels in downstream tasks such as object detection, instance segmentation, and semantic segmentation.  Its robustness is further validated across diverse datasets.

\end{abstract}
\section{Introduction}

 Since its inception, the Vision Transformer (ViT)~\cite{vit} has attracted considerable attention from the research community, primarily owing to its exceptional capability in modeling long-range dependencies.
 However, the self-attention mechanism~\cite{attention}, as the core of  ViT, imposes significant computational overhead, thus constraining its broader applicability. Several strategies have been proposed to alleviate this limitation of self-attention. For instance, methods such as Swin-Transformer~\cite{SwinTransformer, cswin} group tokens for attention, reducing computational costs and enabling the model to focus more on local information. Techniques like PVT~\cite{pvt, pvtv2, cmt, FAT, stvit} down-sample tokens to shrink the size of the QK matrix, thus lowering computational demands while retaining global information. Meanwhile, approaches such as UniFormer~\cite{uniformer, litv2} forgo attention operations in the early stages of visual modeling, opting instead for lightweight convolution. Furthermore, some models~\cite{dynamicvit} enhance computational efficiency by pruning redundant tokens.

\begin{figure}
    \centering
    \includegraphics[width=0.99\textwidth]{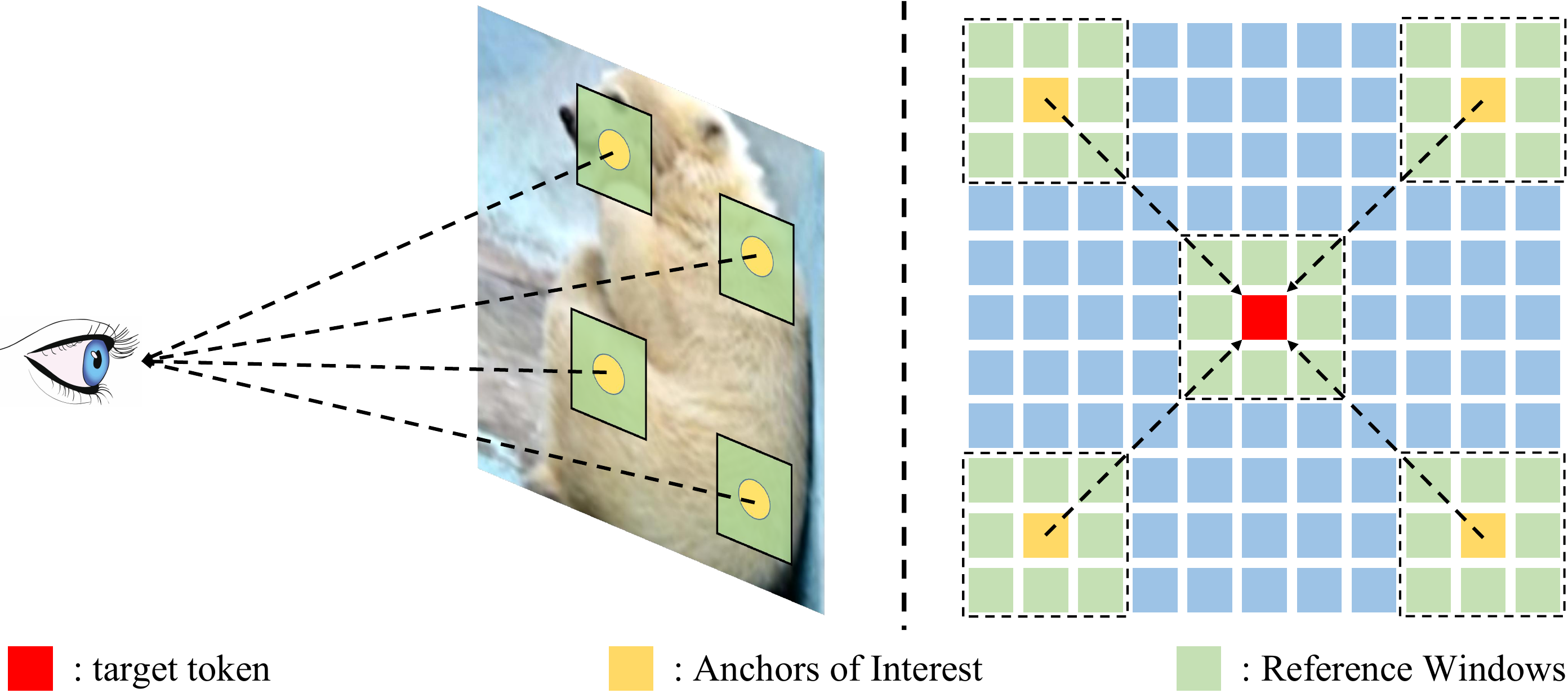}
    \caption{Illustration of sparse scan mechanism in human eye and our $\rm{S}^3\rm{A}$. \textbf{Left:} The human eye exhibits a sparse scan mechanism when observing visual information, focusing only on a few anchors of interest and the local information surrounding these anchors. It doesn't soly model global/local information. \textbf{Right: }Our proposed $\rm{S}^3\rm{A}$ mimics the sparse scan mechanism illustrated in the left figure, focusing on modeling the local information surrounding the Anchors of Interest.}
    \vspace{-5mm}
    \label{fig:help}
\end{figure}

Despite these advancements, the majority of methods primarily focus on reducing the token count in self-attention operations to boost ViT efficiency, often neglecting the manner in which human eyes process visual information. The human visual system operates in a notably less intricate yet highly efficient manner compared to ViT models. Unlike the fine-grained local spatial information modeling in models like Swin~\cite{SwinTransformer}, NAT~\cite{NAT}, LVT~\cite{LVT}, or the indistinct global information modeling seen in models like PVT~\cite{pvt}, PVTv2~\cite{pvtv2}, CMT~\cite{cmt}, human vision employs a sparse scanning mechanism, as substantiated by numerous biological studies~\cite{ss1, ss2, ss3}. As illustrated in Fig~\ref{fig:help}, our eyes swiftly move between points of interest, delving into detailed information processing solely at these anchor points~\cite{ss4, ss5, ss6}. This selective attention mechanism enables the brain to efficiently process essential visual information, rather than being solely focused on local details or vague global information. Given that the human retina's fovea has a fixed size, each shift in the eye's focal point results in a fixed-size receptive field being sensed~\cite{receptive1, receptive2}.


\begin{wrapfigure}{r}{0.5\textwidth}
    \centering
    \includegraphics[width=0.5\textwidth]{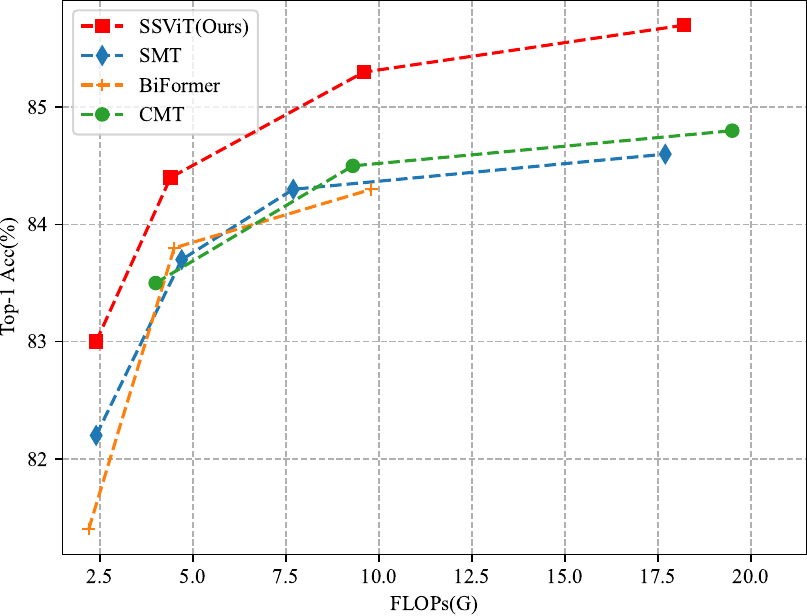}
    \caption{Top-1 accuracy v.s. FLOPs on ImageNet-1K of recent SOTA models. Our SSViT outperforms all the counterparts in all settings.}
    \label{fig:comp}
    \vspace{-4mm}
\end{wrapfigure}


As depicted in Fig~\ref{fig:help}, we introduce a novel Self-Attention mechanism, termed \textbf{S}parse \textbf{S}can \textbf{S}elf-\textbf{A}ttention ($\rm{S}^3\rm{A}$), inspired by the sparse scanning mechanism of the human eye. For each target token, we design a set of uniformly distributed \textbf{A}nchors \textbf{o}f \textbf{I}nterest (AoI). We apply local attention to these AoIs, processing the surrounding visual information and utilizing this local data to update the AoI tokens. The size of each local window remains constant, reflecting the fixed foveal size in human vision. We subsequently aggregate the information from all AoIs to update the target token. The $\rm{S}^3\rm{A}$ modeling approach harmonizes fine-grained local modeling and sparse modeling of interest anchors, closely mirroring the functioning of the human eye. $\rm{S}^3\rm{A}$'s methodology surpasses previous Self-Attention mechanisms, offering a more human-like, efficient, and effective model.

Building on S³A, we develop the Sparse Scan Vision Transformer (SSViT). SSViT effectively replicates the human eye's visual information processing and demonstrates remarkable effectiveness across a spectrum of visual tasks. As demonstrated in Fig~\ref{fig:comp}, SSViT outperforms previous state-of-the-art models in image classification accuracy, achieving 83.0\% top-1 accuracy with a mere 15M parameters and 2.4G FLOPs, without the need for additional training data or supervision. This performance advantage is sustained even when the model scales up, with our SSViT-L achieving 85.7\% top-1 accuracy with only 100M parameters. Beyond classification tasks, SSViT also excels in downstream tasks such as object detection, instance segmentation, and semantic segmentation. The robustness of SSViT is further corroborated by its superior performance across a variety of datasets.

\section{Related works}
\paragraph{Vision Transformers.} 
The Vision Transformer (ViT)\cite{vit} has attracted significant attention since its inception, largely due to its superior performance. Numerous studies\cite{SwinTransformer, cswin, FAT, cloformer} have explored ways to optimize ViT by refining its central operator, Self-Attention, with the dual objective of reducing its quadratic computational complexity and enhancing its performance. 
A range of methods~\cite{cswin, SwinTransformer, biformer} have been proposed to reduce the computational burden of Self-Attention. These techniques limit the region that each token can attend to by grouping tokens. The Swin-Transformer~\cite{SwinTransformer}, for instance, divides all tokens into separate windows and performs Self-Attention operations within these windows. The BiFormer~\cite{biformer}, in contrast, dynamically determines the windows that each token can attend to. 
Additionally, some methods~\cite{pvt, pvtv2, cmt, stvit, globalvit, litv2, regionvit} reduce the number of tokens involved in Self-Attention operations through token downsampling. The PVT~\cite{pvt, pvtv2} employs average pooling for direct downsampling, thus decreasing the number of tokens. The CMT~\cite{cmt} and PVTv2~\cite{pvtv2} supplement token downsampling with convolution to enhance the model's ability to learn local features. The STViT~\cite{stvit} effectively captures global dependencies by sampling super tokens, applying self-attention to them, and subsequently mapping them back to the original token space.
Certain approaches~\cite{SMT, litv2, uniformer} choose to forego the computationally demanding Self-Attention in the early layers of the model, instead employing more efficient convolutions to learn local features. Self-Attention is then deployed in the deeper layers of the model to learn global features. 
While these aforementioned methods exhibit promising results and reduce computational complexity, it is important to note that their operational mechanisms significantly deviate from the functioning of the human eye.

\paragraph{Sparse Scan in Human Eye.}
Sparse scanning, a critical mechanism in human vision, facilitates the efficient processing of visual stimuli in the face of sensory limitations. Neuroimaging investigations have pinpointed key neural structures, such as the superior colliculus, that govern sparse scanning behaviors~\cite{Thompson2005}. Furthermore, microsaccades, minute ocular movements considered a variant of sparse scanning, are instrumental in preserving visual stability and guiding attention~\cite{ss6}. Impairments in sparse scanning are linked to cognitive deficits observed in disorders like ADHD and schizophrenia~\cite{ss1}. 
The dynamism of sparse scanning in relation to task requirements and attentional preferences has also been a focus of research. Evidence suggests that sparse scanning exhibits adaptability to task-specific demands, flexibly distributing visual resources to enhance processing efficiency~\cite{ss5}. This adaptive characteristic underscores the complex interplay between bottom-up sensory inputs and top-down cognitive influences in visual processing. 
In essence, sparse scanning is a fundamental mechanism that impacts not only basic perceptual functions but also more complex cognitive processes~\cite{ss2, ss3, ss4}.
\begin{figure*}
    \centering
    \includegraphics[width=1.00\textwidth]{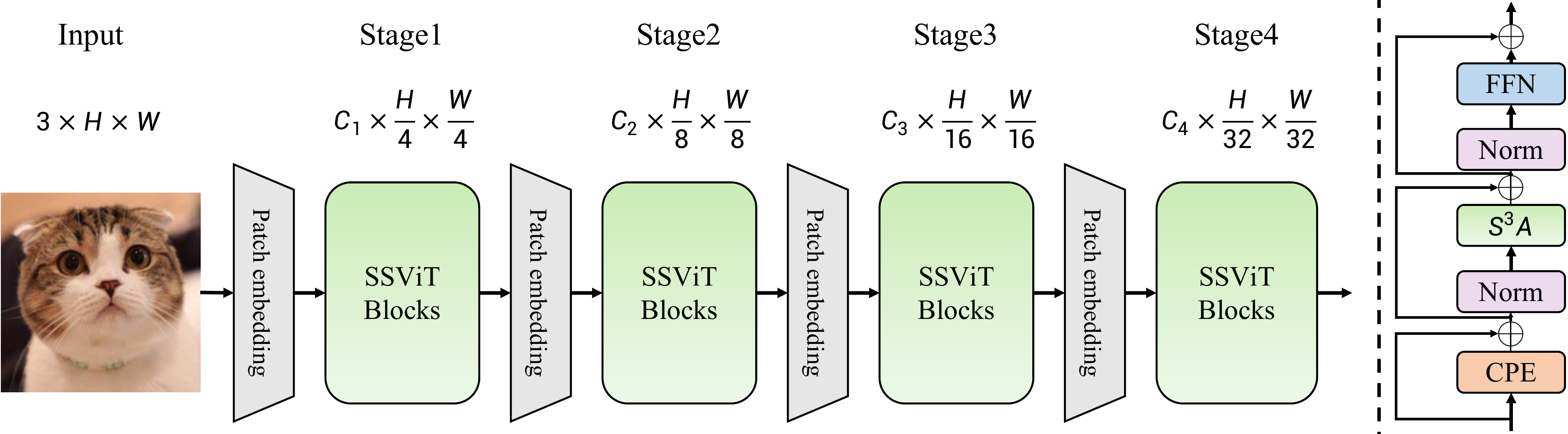}
    \caption{Illustration of the SSViT. SSViT consists of multiple SSViT blocks. A single SSViT block is composed of CPE, ${\rm{S}^3\rm{A}}$, and FFN.}
    \label{fig:main}
    
\end{figure*}

\section{Method}
\subsection{Overall Architecture.}
The overall architecture of Sparse Scan Vision Transformer (SSViT) is shown in Fig.~\ref{fig:main}. To process the input image $x\in \mathbb{R}^{3\times H\times W}$, we feed it into a patch embedding composed of convolutions, obtaining tokens with a shape of $C_1\times \frac{H}{4} \times \frac{W}{4}$. Following the previous hierarchical designs~\cite{SwinTransformer, cloformer, FAT}, we divide SSViT into four stages. The multi-resolution representations brought by hierarchical structures can be utilized for downstream tasks such as object detection and semantic segmentation.

An SSViT block consists of three key components: Conditional Positional Encoding (CPE)~\cite{CPVT}, Sparse Scan Self-Attention ($\rm{S}^3\rm{A}$), and Feed-Forward Network (FFN)~\cite{attention, vit}. The complete SSViT block can be defined as Eq.~\ref{eq:ssvit_block}:
\begin{equation}
\label{eq:ssvit_block}
\centering
\begin{split}
    &X={\rm CPE(}X_{in}{\rm )}+X_{in}, \\
    &Y={\rm S^3A(}{\rm LN(}X{\rm))}+X, \\
    &Z={\rm FFN(}{\rm LN(}Y{\rm ))}+Y.
\end{split}
\end{equation}
For each block, the input tensor $X_{in}\in \mathbb{R}^{C\times H\times W}$ is fed into the CPE to introduce the positional information for each token. After CPE, $\rm{S}^3\rm{A}$ is employed to scan sparse regions of interest for each token. The final FFN is utilized to integrate channel-wise information of tokens.
\begin{figure}
    \centering
    \includegraphics[width=1.0\linewidth]{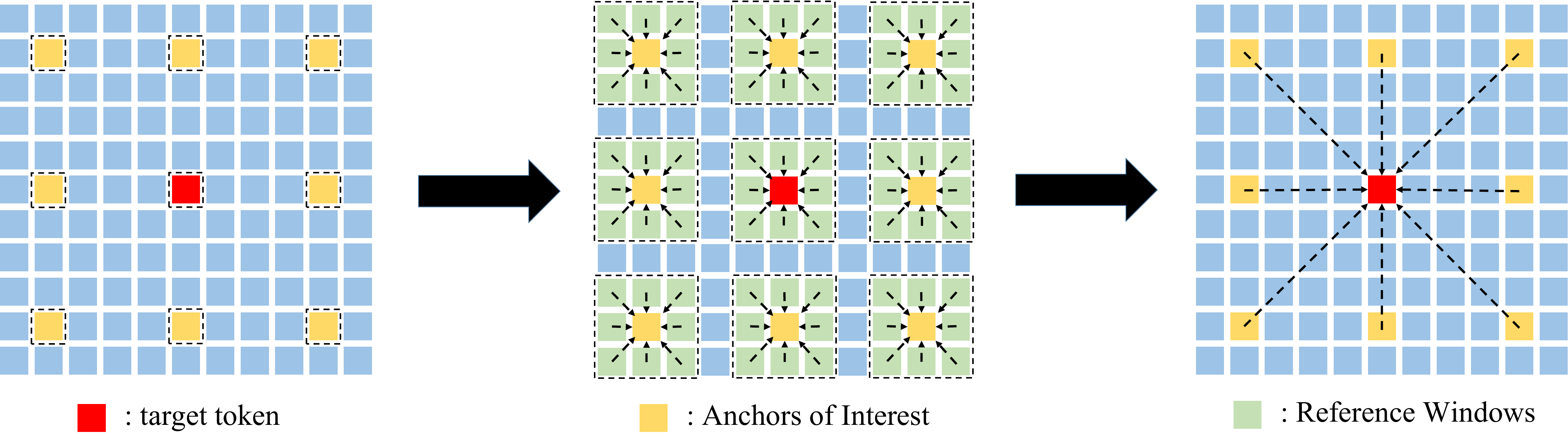}
    \caption{Illustration of Sparse Scan Self-Attention.}
    \label{fig:s3a}
    \vspace{-4mm}
\end{figure}

\subsection{Sparse Scan Self-Attention}
The Sparse Scan Self-Attention ($\rm{S^3A}$) is inspired by the sparse scan mechanism of the human eye in processing visual information. It can be decomposed into three sub-processes. \textbf{Firstly}, the selection of Anchors of Interest (AoI) for each token. \textbf{Secondly}, the extraction of background local information within the reference windows (RWins) determined by the AoIs. \textbf{Finally}, the interaction among AoIs. The whole process can be seen in Fig.~\ref{fig:s3a}.
\paragraph{Anchors of Interests.}Ideally, each token should select its own suitable AoIs based on its characteristics. However, this approach would grant the model excessive freedom, resulting in a cumbersome implementation process and low efficiency. Therefore, we have abandoned this practice and instead opt to artificially define the AoIs for each token. Specifically, assuming the stride of the AoIs is $(S_h, S_w)$, and the number of AoIs chosen for each token is $(N_h, N_w)$. For a token located at position $(i,j)$, its selected AoIs can be represented as Eq.~\ref{eq:AoIs}:
\begin{equation}
\label{eq:AoIs}
\centering
\begin{split}
     \mathbb{A}_{i,j}=&\{X_{m, n}|m=i+k_h \times S_h, n=j+k_w \times S_w \}, \\
    & k_h \in (-\lfloor N_h/2\rfloor, \lfloor N_h/2\rfloor), k_h \in \mathbb{Z}, \\
    & k_w \in (-\lfloor N_w/2\rfloor, \lfloor N_w/2\rfloor), k_w \in \mathbb{Z}. \\
\end{split}
\end{equation}
where $X$ is the input feature map, and $\mathbb{A}_{i,j}$ is the set of AoIs for $X_{ij}$. For the sake of simplicity, in Eq.~\ref{eq:AoIs}, we do not consider the situation of boundary points. In practice, when dealing with boundary points, there is a certain offset in the range of $k_h$ and $k_w$ to ensure that all AoIs remain within the boundaries of the feature map.

\paragraph{Background Local Information.}As shown in Fig.~\ref{fig:help}, when the human eye observes certain anchors, it also processes the surrounding background information. Based on this consideration, for each anchor point in the AoI from the previous step, we select its corresponding reference window (RWin), which is the background of the anchor point. Consistent with the previous definition, we assume the size of a single RWin is $(W_h, W_w)$. Since the size of the fovea on the human eye's retina remains constant, the receptive field perceived by the human eye is the same for each AoI. Translated into model design, this means that each AoI has a RWin of the same size. For an AoI located at position $(m, n)$, its RWin is defined as Eq.~\ref{eq:RWin}:
\begin{equation}
\label{eq:RWin}
\centering
\begin{split}
      \mathbb{R}_{m,n}&=\{X_{p,q}|p=m+r_h, q=n+r_w \}, \\
      &r_h \in  [-\lfloor W_h/2 \rfloor, \lfloor W_h/2 \rfloor], r_h \in \mathbb{Z}, \\
      &r_w \in  [-\lfloor W_w/2 \rfloor, \lfloor W_w/2 \rfloor], r_w \in \mathbb{Z}, \\
      &X_{m,n}\in \mathbb{A}_{i,j}
\end{split}
\end{equation}
Where $\mathbb{R}_{m,n}$ is the set of tokens in RWin of $X_{m,n}$. Similar to the definition of AoI, in Eq.~\ref{eq:RWin}, we omit the situation of boundary points. For each AoI, within its determined RWin, we utilize Self-Attention to update the AoI. For an AoI with the position of $(m, n)$, this process can be represented as Eq.~\ref{eq:attn}:
\begin{equation}
\label{eq:attn}
\centering
\begin{split}
      X^*_{m,n}&={\rm Attn}(W_q X_{m,n}, W_k\mathbb{R}_{m, n}, W_v\mathbb{R}_{m, n}),\\
      \mathbb{A}&^*_{i,j}=\{X^*_{m,n}|X_{m,n}\in \mathbb{A}_{i,j} \},
\end{split}
\end{equation}
where ${\rm Attn}(q, k, v)$ denotes the standard Self-Attention operation. $W_q, W_k, W_v$ are learnable matrices. $X^*_{m,n}$ is the updated $X_{m,n}$. $\mathbb{A}^*_{i,j}$ is the set of updated AoIs for $X_{i,j}$.

\paragraph{Interaction among Anchors.} In practice, the human eye does not process each anchor independently. Instead, it performs interactive modeling of the information observed in each anchor, thereby inferring deep semantic information from the image. We also model this process by Self-Attention. For the target token $X_{i,j}$, after all its AoIs have been updated, we utilize $\mathbb{A}_{i,j}$ and $\mathbb{A}^*_{i,j}$ to update $X_{i,j}$, as shown specifically in Eq.~\ref{eq:attn2}:
\begin{equation}
\label{eq:attn2}
\centering
\begin{split}
     X_{i,j}={\rm Attn}(W_qX_{i,j}, W_k\mathbb{A}_{i,j}, \mathbb{A}^*_{i,j}).
\end{split}
\end{equation}
The above three-step process constitutes the complete ${\rm S^3A}$. After the completion of ${\rm S^3A}$, to further enhance the model's ability to capture local information, we employ a local context enhancement module to model local information:
\begin{equation}
\label{eq:lce}
\centering
\begin{split}
     X = {\rm S^3A}(X)+{\rm LCE}(W_vX),
\end{split}
\end{equation}
where ${\rm LCE}$ is a simple depth-wise convolution. It is worth noting that although ${\rm S^3A}$ performs two rounds of Self-Attention calculations, in practice, the projection of $q$, $k$, and $v$ is completed in a single operation. During the two self-attention computations (Eq.~\ref{eq:attn} and Eq.~\ref{eq:attn2}), $q$ and $k$ are reused, thus no additional computational or parameter overhead is introduced.

\section{Experiments}
\label{sec:exp}
We conducted experiments on a wide range of vision tasks, including image classification on ImageNet-1k~\cite{imagenet}, object detection and instance segmentation on COCO~\cite{coco}, and semantic segmentation on ADE20K~\cite{ade20k}. We also evaluate the SSViT's robustness on ImageNet-v2~\cite{imagenetv2}, ImageNet-A~\cite{imagenet-a}, ImageNet-R~\cite{imagenet-r}. All models can be trained with 8 A100 80G GPUs.
\subsection{ImageNet Classification}
\paragraph{Settings.}We train our models from scratch on ImageNet-1k~\cite{imagenet}. For a fair comparison, we adopt the same training strategy as in~\cite{SwinTransformer}, with classification loss serving as the sole supervision. The maximum rates for increasing stochastic depth~\cite{droppath} are set to 0.1, 0.15, 0.4, and 0.5 for SSViT-T, SSViT-S, SSViT-B, and SSViT-L, respectively.
\paragraph{Comparison with SOTA.} We benchmark our SSViT against numerous state-of-the-art models, with results presented in Tab.\ref{tab:ImageNet}. SSViT consistently outperforms preceding models across all scales. Notably, SSViT-T attains a Top1-accuracy of \textbf{83.0\%} with a mere 15M parameters and \textbf{2.4G} FLOPs, exceeding the previous state-of-the-art (SMT\cite{SMT}) by \textbf{0.8\%}. For larger models, SSViT-L achieves a Top1-accuracy of \textbf{85.7\%} with \textbf{100M} parameters and \textbf{18.2G} FLOPs.

\begin{table*}[ht]
    \centering
    \setlength{\tabcolsep}{1.8mm}
    \subfloat{
    \scalebox{0.75}{
    \begin{tabular}{c|c|c c|c}
        \toprule[1pt]
        \makecell{Cost} & Model & \makecell{Parmas\\(M)} & \makecell{FLOPs\\(G)} & \makecell{Top1-acc\\(\%)}\\
        \midrule[0.5pt]
        \multirow{13}{*}{\rotatebox{90}{\makecell{tiny model\\$\sim 2.5$G}}} 
        &QuadTree-B-b1~\cite{quadtree} & 14 & 2.3 & 80.0 \\
        &RegionViT-T~\cite{regionvit} & 14 & 2.4 & 80.4 \\
        &MPViT-XS~\cite{mpvit} & 11 & 2.9 & 80.9 \\
        &VAN-B1~\cite{VAN} & 14 & 2.5 & 81.1 \\
        &BiFormer-T~\cite{biformer} & 13 & 2.2 & 81.4 \\
        &Conv2Former-N~\cite{conv2former} & 15 & 2.2 & 81.5 \\
        &CrossFormer-T~\cite{crossformer} & 28 & 2.9 & 81.5 \\
        &NAT-M~\cite{NAT} & 20 & 2.7 & 81.8 \\
        &FAT-B2~\cite{FAT} & 14 & 2.0 & 81.9 \\
        &QnA-T~\cite{QnA} & 16 & 2.5 & 82.0 \\
        &GC-ViT-XT~\cite{globalvit} & 20 & 2.6 & 82.0 \\
        &SMT-T~\cite{SMT} & 12 & 2.4 & 82.2 \\
        &\cellcolor{gray!30}SSViT-T & \cellcolor{gray!30}15 & \cellcolor{gray!30}2.4 & \cellcolor{gray!30}\textbf{83.0} \\
        \midrule[0.5pt]
        \multirow{15}{*}{\rotatebox{90}{\makecell{small model\\$\sim 4.5$G}}} 
        &Swin-T~\cite{SwinTransformer} & 29 & 4.5 & 81.3 \\
        &CrossViT-15~\cite{crossvit} & 27 & 5.8 & 81.5 \\
        &RVT-S~\cite{RVT} & 23 & 4.7 & 81.9 \\
        &ConvNeXt-T~\cite{convnext} & 29 & 4.5 & 82.1 \\
        &Focal-T~\cite{focal} & 29 & 4.9 & 82.2 \\
        &MPViT-S~\cite{mpvit} & 23 & 4.7 & 83.0 \\
        &SG-Former-S~\cite{sgformer} & 23 & 4.8 & 83.2 \\
        &Ortho-S~\cite{Ortho} & 24 & 4.5 & 83.4 \\
        &InternImage-T~\cite{internimage} & 30 & 5.0 & 83.5 \\
        &GC-ViT-T~\cite{globalvit} & 28 & 4.7 & 83.5 \\
        &CMT-S~\cite{cmt} & 25 & 4.0 & 83.5 \\
        &FAT-B3~\cite{FAT} & 29 & 4.4 & 83.6 \\
        &SMT-S~\cite{SMT} & 20 & 4.8 & 83.7 \\
        &BiFormer-S~\cite{biformer} & 26 & 4.5 & 83.8 \\
        &\cellcolor{gray!30}SSViT-S & \cellcolor{gray!30}27 & \cellcolor{gray!30}4.4 & \cellcolor{gray!30}\textbf{84.4} \\
        \bottomrule[1pt]
    \end{tabular}}}
    \hfill
    \setlength{\tabcolsep}{1.8mm}
    \subfloat{
    \scalebox{0.75}{
    \begin{tabular}{c|c|c c|c}
        \toprule[1pt]
        \makecell{Cost} & Model & \makecell{Parmas\\(M)} & \makecell{FLOPs\\(G)} & \makecell{Top1-acc\\(\%)}\\
        \midrule[0.5pt]
        \multirow{14}{*}{\rotatebox{90}{\makecell{base model\\$\sim 9.0$G}}}
        &ConvNeXt-S~\cite{convnext} & 50 & 8.7 & 83.1 \\
        &CrossFormer-B~\cite{crossformer} & 52 & 9.2 & 83.4 \\
        &NAT-S~\cite{NAT} & 51 & 7.8 & 83.7 \\
        &Quadtree-B-b4~\cite{quadtree} & 64 & 11.5 & 84.0 \\
        &ScaleViT-B~\cite{ScalableViT} & 81 & 8.6 & 84.1 \\
        &MOAT-1~\cite{MOAT} & 42 & 9.1 & 84.2 \\
        &InternImage-S~\cite{internimage} & 50 & 8.0 & 84.2 \\
        &DaViT-S~\cite{davit} & 50 & 8.8 & 84.2 \\
        &BiFormer-B~\cite{biformer} & 57 & 9.8 & 84.3 \\
        &MViTv2-B~\cite{mvitv2} & 52 & 10.2 & 84.4 \\
        &CMT-B~\cite{cmt} & 46 & 9.3 & 84.5 \\
        &iFormer-B~\cite{iformer} & 48 & 9.4 & 84.6 \\
        &STViT-B~\cite{stvit} & 52 & 9.9 & 84.8 \\
        &\cellcolor{gray!30}SSViT-B & \cellcolor{gray!30}57 & \cellcolor{gray!30}9.6 & \cellcolor{gray!30}\textbf{85.3} \\
        
        \midrule[0.5pt]
        \multirow{14}{*}{\rotatebox{90}{\makecell{large model\\$\sim 18.0$G}}}
        &DeiT-B~\cite{deit} & 86 & 17.5 & 81.8 \\
        &LITv2~\cite{litv2} & 87 & 13.2 & 83.6 \\
        &CrossFormer-L~\cite{crossformer} & 92 & 16.1 & 84.0 \\
        &Ortho-L~\cite{Ortho} & 88 & 15.4 & 84.2 \\
        &CSwin-B~\cite{cswin} & 78 & 15.0 & 84.2 \\
        &SMT-L~\cite{SMT} & 81 & 17.7 & 84.6 \\
        &DaViT-B~\cite{davit} & 88 & 15.5 & 84.6 \\
        &SG-Former-B~\cite{sgformer} & 78 & 15.6 & 84.7 \\
        &iFormer-L~\cite{iformer} & 87 & 14.0 & 84.8 \\
        &CMT-L~\cite{cmt} & 75 & 19.5 & 84.8 \\
        &InterImage-B~\cite{internimage} & 97 & 16.0 & 84.9 \\
        &MaxViT-B~\cite{maxvit} & 120 & 23.4 & 84.9 \\
        &GC-ViT-B~\cite{globalvit} & 90 & 14.8 & 85.0 \\
        &\cellcolor{gray!30}SSViT-L & \cellcolor{gray!30}100 & \cellcolor{gray!30}18.2 & \cellcolor{gray!30}\textbf{85.7} \\
        \bottomrule[1pt]
    \end{tabular}}}
    
    \caption{Comparison with the state-of-the-art on ImageNet-1K classification.}
    \vspace{-3mm}
    \label{tab:ImageNet}
\end{table*}

\paragraph{Strict Comparison with General/Efficient Models.} To guarantee a fair comparison, we select two baselines: the general-purpose backbone, Swin-Transformer~\cite{SwinTransformer}, and the efficiency-oriented backbone, FasterViT~\cite{fastervit}. We compare these with SSViT. In the comparison models (SS-Swin and SS-FasterViT), we merely substitute the attention mechanism in the original Swin-Transformer and FasterViT with ${\rm S^3A}$, without introducing any other modifications (such as CPE, Conv Stem, etc.). As shown in Tab.~\ref{tab:ImageNet_baselines}, the simple replacement of the attention mechanism with ${\rm S^3A}$ yields significant advantages in both performance and efficiency. Specifically, SS-Swin achieves or even surpasses a \textbf{2.0\%} improvement over Swin across all model sizes. Meanwhile, SS-FasterViT attains higher accuracy than FasterViT while utilizing fewer parameters.


\begin{table*}[ht]
    \centering
    \setlength{\tabcolsep}{1mm}
    \subfloat{
    \scalebox{0.82}{
    \begin{tabular}{c|c c c|c}
    \toprule[1pt]
         Model & \makecell{Parmas\\(M)} & \makecell{FLOPs\\(G)} &\makecell{Throughput\\(imgs/s)} &\makecell{Top1-acc\\(\%)}\\
         \midrule[0.5pt]
         Swin-T~\cite{SwinTransformer} & 29 & 4.5 & 1723 & 81.3 \\
         \rowcolor{gray!30}SS-Swin-T & 29 & 4.8 & 1282 & \textbf{83.8}(\textcolor{red}{+2.5}) \\
         \midrule[0.5pt]
         Swin-S~\cite{SwinTransformer} & 50 & 8.8 & 1062 & 83.0 \\
         \rowcolor{gray!30}SS-Swin-S & 50 & 8.9 & 724 &\textbf{84.7}(\textcolor{red}{+1.7}) \\
         \midrule[0.5pt]
         Swin-B~\cite{SwinTransformer} & 88 & 15.4 & 798 & 83.3 \\
         \rowcolor{gray!30}SS-Swin-B & 88 & 15.7 & 538 & \textbf{85.1}(\textcolor{red}{+1.8}) \\
        \bottomrule[1pt]      
    \end{tabular}}}
    \hfill
    \centering
    \subfloat{
    \scalebox{0.82}{
    \begin{tabular}{c|c c c|c}
        \toprule[1pt]
         Model & \makecell{Parmas\\(M)} & \makecell{FLOPs\\(G)} &\makecell{Throughput\\(imgs/s)} &\makecell{Top1-acc\\(\%)}\\
         \midrule[0.5pt]
         FasterViT-0~\cite{fastervit} & 31 & 3.3 & 3551 & 82.1 \\
         \rowcolor{gray!30}SS-FasterViT-0 & 25 & 3.4 & 3021 & \textbf{82.8}(\textcolor{red}{+0.7}) \\
         \midrule[0.5pt]
         FasterViT-1~\cite{fastervit} & 53 & 5.3 & 2619 & 83.2 \\
         \rowcolor{gray!30}SS-FasterViT-1 & 41 & 5.4 & 2282 & \textbf{83.7}(\textcolor{red}{+0.5}) \\
         \midrule[0.5pt]
         FasterViT-2~\cite{fastervit} & 76 & 8.7 & 1988 & 84.2 \\
         \rowcolor{gray!30}SS-FasterViT-2 & 58 & 8.8 & 1783 & \textbf{84.6}(\textcolor{red}{+0.4}) \\
         \bottomrule[1pt]
    \end{tabular}}}
    
    \caption{Strict comparison with the baslines on ImageNet-1K classfication. The speed of the models are measured on A100 GPU with the batch size of 64.}
    \label{tab:ImageNet_baselines}
\end{table*}

\subsection{Object Detection and Instance Segmentation}

\paragraph{Settings.} We utilize MMDetection~\cite{mmdetection} to implement Mask-RCNN~\cite{maskrcnn}, Cascade Mask R-CNN~\cite{cai18cascadercnn}, and RetinaNet~\cite{retinanet} for evaluating our SSViT. For Mask R-CNN and Cascade Mask R-CNN, we adhere to the commonly used ``$3\times +$ MS'' setting, and for Mask R-CNN and RetinaNet, we apply the ``$1\times$'' setting. Following~\cite{SwinTransformer}, during training, we resize the images such that the shorter side is 800 pixels while keeping the longer side within 1333 pixels. We employ the AdamW optimizer for model optimization.


\begin{table}[ht]
    \centering
    \setlength{\tabcolsep}{0.1mm}
    \subfloat{
    \scalebox{0.71}{
    \begin{tabular}{c|c c|c c c c c c}
        \toprule[1pt]
         \multirow{2}{*}{Backbone} & \multirow{2}{*}{\makecell{Params\\(M)}} & \multirow{2}{*}{\makecell{FLOPs\\(G)}} & \multicolumn{6}{c}{Mask R-CNN $3\times$+MS}\\
          & & & $AP^b$ & $AP^b_{50}$ & $AP^b_{75}$ & $AP^m$ & $AP^m_{50}$ & $AP^m_{75}$\\
          \midrule[0.5pt]
          Focal-T~\cite{focal} & 49 & 291 & 47.2 & 69.4 & 51.9 & 42.7 & 66.5 & 45.9 \\
          NAT-T~\cite{NAT} & 48 & 258 & 47.8 & 69.0 & 52.6 & 42.6 & 66.0 & 45.9 \\
          GC-ViT-T~\cite{globalvit} & 48 & 291 & 47.9 & 70.1 & 52.8 & 43.2 & 67.0 & 46.7 \\
          MPViT-S~\cite{mpvit} & 43 & 268 & 48.4 & 70.5 & 52.6 & 43.9 & 67.6 & 47.5 \\
          SMT-S~\cite{SMT} & 40 & 265 & 49.0 & 70.1 & 53.4 & 43.4 & 67.3 & 46.7\\
          CSWin-T~\cite{cswin} & 42 & 279 & 49.0 & 70.7 & 53.7 & 43.6 & 67.9 & 46.6\\
          InternImage-T~\cite{internimage} & 49 & 270 & 49.1 & 70.4 & 54.1 & 43.7 & 67.3 & 47.3 \\
          \rowcolor{gray!30}SSViT-S & 46 & 266 & \textbf{51.2} & \textbf{72.0} & \textbf{56.0} & \textbf{45.4} & \textbf{69.7} & \textbf{49.0}\\
          \midrule[0.5pt]
          ConvNeXt-S~\cite{convnext} & 70 & 348 & 47.9 & 70.0 & 52.7 & 42.9 & 66.9 & 46.2 \\
          NAT-S~\cite{NAT} & 70 & 330 & 48.4 & 69.8 & 53.2 & 43.2 & 66.9 & 46.4 \\
          Swin-S~\cite{SwinTransformer} & 69 & 359 & 48.5 & 70.2 & 53.5 & 43.3 & 67.3 & 46.6 \\
          InternImage-S~\cite{internimage} & 69 & 340 & 49.7 & 71.1 & 54.5 & 44.5 & 68.5 & 47.8 \\
          SMT-B~\cite{SMT} & 52 & 328 & 49.8 & 71.0 & 54.4 & 44.0 & 68.0 & 47.3\\
          CSWin-S~\cite{cswin} & 54 & 342 & 50.0 & 71.3 & 54.7 & 44.5 & 68.4 & 47.7 \\
          \rowcolor{gray!30}SSViT-B & 76 & 382 & \textbf{52.6} & \textbf{73.2} & \textbf{57.7} & \textbf{46.4} & \textbf{70.9} & \textbf{50.3}\\
          \bottomrule
    \end{tabular}}}
    \hfill
    \subfloat{
    \scalebox{0.71}{
    \begin{tabular}{c|c c|c c c c c c}
        \toprule[1pt]
         \multirow{2}{*}{Backbone} & \multirow{2}{*}{\makecell{Params\\(M)}} & \multirow{2}{*}{\makecell{FLOPs\\(G)}} & \multicolumn{6}{c}{Cascade Mask R-CNN $3\times$+MS}\\
          & & & $AP^b$ & $AP^b_{50}$ & $AP^b_{75}$ & $AP^m$ & $AP^m_{50}$ & $AP^m_{75}$\\
          \midrule[0.5pt]
          NAT-T~\cite{NAT} & 85 & 737 & 51.4 & 70.0 & 55.9 & 44.5 & 67.6 & 47.9 \\
          GC-ViT-T~\cite{globalvit} & 85 & 770 & 51.6 & 70.4 & 56.1 & 44.6 & 67.8 & 48.3 \\
          SMT-S~\cite{SMT} & 78 & 744 & 51.9 & 70.5 & 56.3 & 44.7 & 67.8 & 48.6 \\
          UniFormer-S~\cite{uniformer} & 79 & 747 & 52.1 & 71.1 & 56.6 & 45.2 & 68.3 & 48.9 \\
          Ortho-S~\cite{Ortho} & 81 & 755 & 52.3 & 71.3 & 56.8 & 45.3 & 68.6 & 49.2 \\
          HorNet-T~\cite{hornet} & 80 & 728 & 52.4 & 71.6 & 56.8 & 45.6 & 69.1 & 49.6 \\
          CSWin-T~\cite{cswin} & 80 & 757 & 52.5 & 71.5 & 57.1 & 45.3 & 68.8 & 48.9 \\
          \rowcolor{gray!30}SSViT-S & 84 & 745 & \textbf{53.8} & \textbf{72.4} & \textbf{58.1} & \textbf{46.6} & \textbf{70.1} & \textbf{50.4}\\
          \midrule[0.5pt]
          Swin-S~\cite{SwinTransformer} & 107 & 838 & 51.9 & 70.7 & 56.3 & 45.0 & 68.2 & 48.8 \\
          NAT-S~\cite{NAT} & 108 & 809 & 51.9 & 70.4 & 56.2 & 44.9 & 68.2 & 48.6 \\
          GC-ViT-S~\cite{globalvit} & 108 & 866 & 52.4 & 71.0 & 57.1 & 45.4 & 68.5 & 49.3\\
          DAT-S~\cite{dat} & 107 & 857 & 52.7 & 71.7 & 57.2 & 45.5 & 69.1 & 49.3 \\
          CSWin-S~\cite{cswin} & 92 & 820 & 53.7 & 72.2 & 58.4 & 46.4 & 69.6 & 50.6 \\
          UniFormer-B~\cite{uniformer} & 107 & 878 & 53.8 & 72.8 & 58.5 & 46.4 & 69.9 & 50.4 \\
          \rowcolor{gray!30}SSViT-B & 114 & 861 & \textbf{54.9} & \textbf{73.7} & \textbf{59.7} & \textbf{47.6} & \textbf{71.6} & \textbf{51.5}  \\
          \bottomrule
    \end{tabular}}}
    \caption{Comparison with other backbones using "$3\times+\mathrm{MS}$`` schedule on COCO.}
    \label{tab:COCO3x}
    \vspace{-1mm}
\end{table}

\paragraph{Results.} 
Tab.~\ref{tab:COCO3x} and Tab.~\ref{tab:COCO1x} present the performance of SSViT across different detection frameworks. The results highlight that SSViT consistently outperforms its counterparts in all comparisons. Under the ``$3\times+$MS'' schedule, SSViT surpasses the recent SMT, achieving a \textbf{+2.2} box AP and \textbf{+2.0} mask AP improvement with the Mask R-CNN framework. For Cascade Mask R-CNN, SSViT still maintains a significant performance edge over SMT. Regarding the ``$1\times$'' schedule, SSViT exhibits remarkable performance. Specifically, SSViT-S attains an improvement of \textbf{+2.2} box AP and \textbf{+1.5} mask AP over InternImage-T within the Mask-RCNN framework.


\begin{table*}[!ht]
    \setlength{\tabcolsep}{0.52mm}
    \centering
    \scalebox{0.78}{
    \begin{tabular}{c|c c|c c c c c c|c c|c c c c c c}
        \toprule[1pt]
        \multirow{2}{*}{Backbone} & \multirow{2}{*}{\makecell{Params\\(M)}} & \multirow{2}{*}{\makecell{FLOPs\\(G)}} & \multicolumn{6}{c|}{Mask R-CNN $1\times$} & \multirow{2}{*}{\makecell{Params\\(M)}} & \multirow{2}{*}{\makecell{FLOPs\\(G)}} & \multicolumn{6}{c}{RetinaNet $1\times$}\\
         & & & $AP^b$ & $AP^b_{50}$ & $AP^b_{75}$ & $AP^m$ & $AP^m_{50}$ & $AP^m_{75}$ & & & $AP^b$ & $AP^b_{50}$ & $AP^b_{75}$ & $AP^b_S$ & $AP^b_{M}$ & $AP^b_{L}$ \\
         \midrule[0.5pt]
        PVT-T~\cite{pvt} & 33 & 240 & 39.8 & 62.2 & 43.0 & 37.4 & 59.3 & 39.9 & 23 & 221 & 39.4 & 59.8 & 42.0 & 25.5 & 42.0 & 52.1 \\
        PVTv2-B1~\cite{pvtv2} & 33 & 243 & 41.8 & 54.3 & 45.9 & 38.8 & 61.2 & 41.6 & 23 & 225 & 41.2 & 61.9 & 43.9 & 25.4 & 44.5 & 54.3 \\
        MPViT-XS~\cite{mpvit} & 30 & 231 & 44.2 & 66.7 & 48.4 & 40.4 & 63.4 & 43.4 & 20 & 211 & 43.8 & 65.0 & 47.1 & 28.1 & 47.6 & 56.5 \\
        \rowcolor{gray!30}SSViT-T & 34 & 223 & \textbf{47.3} & \textbf{69.1} & \textbf{51.7} & \textbf{42.6} & \textbf{66.2} & \textbf{45.8} & 24 & 205 & \textbf{45.6} & \textbf{66.5} & \textbf{49.3} & \textbf{28.6} & \textbf{50.1} & \textbf{60.5} \\
        \midrule[0.5pt]
        CMT-S~\cite{cmt} & 45 & 249 & 44.6 & 66.8 & 48.9 & 40.7 & 63.9 & 43.4 & 44 & 231 & 44.3 & 65.5 & 47.5 & 27.1 & 48.3 & 59.1 \\
        ScalableViT-S~\cite{ScalableViT} & 46 & 256 & 45.8 & 67.6 & 50.0 & 41.7 & 64.7 & 44.8 & 36 & 238 & 45.2 & 66.5 & 48.4 & 29.2 & 49.1 & 60.3 \\
        InternImage-T~\cite{internimage} & 49 & 270 & 47.2 & 69.0 & 52.1  & 42.5 & 66.1 & 45.8 & -- & -- & -- & -- & -- & -- & -- & -- \\
        STViT-S~\cite{stvit} & 44 & 252 & 47.6 & 70.0 & 52.3 & 43.1 & 66.8 & 46.5 & -- & -- & -- & -- & -- & -- & -- & -- \\
        SMT-S~\cite{SMT} & 40 & 265 & 47.8 & 69.5 & 52.1 & 43.0 & 66.6 & 46.1 & -- & -- & -- & -- & -- & -- & -- & -- \\
        BiFormer-S~\cite{biformer} & -- & -- & 47.8 & 69.8 & 52.3 & 43.2 & 66.8 & 46.5 & -- & -- & 45.9 & 66.9 & 49.4 & 30.2 & 49.6 & 61.7 \\
        \rowcolor{gray!30}SSViT-S & 46 & 266 & \textbf{49.4} & \textbf{70.8} & \textbf{54.1} & \textbf{44.0} & \textbf{67.7} & \textbf{47.3} & 36 & 248 & \textbf{47.5} & \textbf{68.6} & \textbf{50.8} & \textbf{30.1} & \textbf{52.2} & \textbf{63.3} \\
        \midrule[0.5pt]
        Swin-S~\cite{SwinTransformer} & 69 & 359 & 45.7 & 67.9 & 50.4 & 41.1 & 64.9 & 44.2 & 60 & 339 & 44.5 & 66.1 & 47.4 & 29.8 & 48.5 & 59.1 \\
        ScalableViT-B~\cite{ScalableViT} &95 & 349 & 46.8 & 68.7 & 51.5 & 42.5 & 65.8 & 45.9 & 85 & 330 & 45.8 & 67.3 & 49.2 & 29.9 & 49.5 & 61.0 \\
        InternImage-S~\cite{internimage} & 69 & 340 & 47.8 & 69.8 & 52.8 & 43.3 & 67.1 & 46.7 & -- & -- & -- & -- & -- & -- & -- & -- \\
        CSWin-S~\cite{cswin} & 54 & 342 & 47.9 & 70.1 & 52.6 & 43.2 & 67.1 & 46.2 & -- & -- & -- & -- & -- & -- & -- & -- \\
        BiFormer-B~\cite{biformer} & -- & -- & 48.6 & 70.5 & 53.8 & 43.7 & 67.6 & 47.1 & -- & -- & 47.1 & 68.5 & 50.4 & 31.3 & 50.8 & 62.6 \\
        \rowcolor{gray!30}SSViT-B & 76 & 382 & \textbf{51.0} & \textbf{72.5} & \textbf{55.8} & \textbf{45.4} & \textbf{69.7} & \textbf{48.9} & 66 & 363 & \textbf{49.0} & \textbf{70.2} & \textbf{52.9} & \textbf{32.4} & \textbf{53.4} & \textbf{64.8} \\
        \midrule[0.5pt]
        Swin-B~\cite{SwinTransformer} & 107 & 496 & 46.9 & 69.2 & 51.6 & 42.3 & 66.0 & 45.5 & 98 & 477 & 45.0 & 66.4 & 48.3 & 28.4 & 49.1 & 60.6 \\
        Focal-B~\cite{focal} & 110 & 533 & 47.8 & 70.2 & 52.5 & 43.2 & 67.3 & 46.5 & 101 & 514 & 46.3 & 68.0 & 49.8 & 31.7 & 50.4 & 60.8 \\
        MPViT-B~\cite{mpvit} & 95 & 503 & 48.2 & 70.0 & 52.9 & 43.5 & 67.1 & 46.8 & 85 & 482 & 47.0 & 68.4 & 50.8 & 29.4 & 51.3 & 61.5 \\
        CSwin-B~\cite{cswin} & 97 & 526 & 48.7 & 70.4 & 53.9 & 43.9 & 67.8 & 47.3 & -- & -- & -- & -- & -- & -- & -- & -- \\
        InternImage-B~\cite{internimage} & 115 & 501 & 48.8 & 70.9 & 54.0 & 44.0 & 67.8 & 47.4 & -- & -- & -- & -- & -- & -- & -- & -- \\
        \rowcolor{gray!30}SSViT-L & 119 & 572 & \textbf{51.6} & \textbf{72.9} & \textbf{56.6} & \textbf{46.0} & \textbf{70.1} & \textbf{49.8} & 109 & 553 & \textbf{50.0} & \textbf{71.4} & \textbf{53.8} & \textbf{33.2} & \textbf{54.6} & \textbf{65.0}\\
        \bottomrule[1pt]
    \end{tabular}}
    \caption{Comparison to other backbones using "$1\times$`` schedule on COCO.}
    \label{tab:COCO1x}
\end{table*}

\subsection{Semantic Segmentation}
\paragraph{Settings.} We utilize Semantic FPN~\cite{semanticfpn} and UperNet~\cite{upernet} to assess SSViT's performance, implementing these frameworks via MMSegmentation~\cite{mmsegmentation}. We mirror PVT's~\cite{pvt} training settings for Semantic FPN, training the model for 80k iterations. All models use an input resolution of $512\times 512$, and during testing, the image's shorter side is resized to 512 pixels. UperNet is trained for 160K iterations, following Swin's~\cite{SwinTransformer} settings. We employ the AdamW optimizer with a weight decay of 0.01, including a 1500 iteration warm-up.

\begin{table}[]
    \centering
    \setlength{\tabcolsep}{0.9mm}
    \subfloat{
    \scalebox{0.84}{
    \begin{tabular}{c|c c|c}
         \toprule[1pt]
         \multicolumn{4}{c}{Semantic FPN}\\
         \midrule[0.5pt]
         Backbone & Params(M) & FLOPs(G) & mIoU(\%)\\
         \midrule[0.5pt]
         PVTv2-B1~\cite{pvtv2}& 18 & 34 & 42.5 \\
         FAT-B2~\cite{FAT} & 17 & 32 & 45.4\\
         EdgeViT-S~\cite{edgevit}& 17 & 32 & 45.9\\
         \rowcolor{gray!30}SSViT-T& 18 & 35 & \textbf{46.8}\\
         \midrule[0.5pt]
         DAT-T~\cite{dat}& 32 & 198 & 42.6 \\
         CSWin-T~\cite{cswin}& 26 & 202 & 48.2 \\
         Shuted-S~\cite{shunted}& 26 & 183 & 48.2 \\
         FAT-B3~\cite{FAT} & 33 & 179 & 48.9 \\
         \rowcolor{gray!30}SSViT-S& 30 & 184 & \textbf{49.6}\\
         \midrule[0.5pt]
         DAT-S~\cite{dat}& 53 & 320 & 46.1\\
         RegionViT-B+~\cite{regionvit} & 77 & 459 & 47.5 \\
         UniFormer-B~\cite{uniformer}& 54 & 350 & 47.7 \\
         CSWin-S~\cite{cswin}& 39 & 271 & 49.2 \\
         \rowcolor{gray!30}SSViT-B& 60 & 303 & \textbf{51.0} \\
         \midrule[0.5pt]
         DAT-B~\cite{dat}& 92 & 481 & 47.0 \\
         CrossFormer-L~\cite{crossformer}& 95 & 497 & 48.7 \\
         CSWin-B~\cite{cswin}& 81 & 464 & 49.9 \\
         \rowcolor{gray!30}SSViT-L& 103 & 497 & \textbf{51.5} \\
         \bottomrule[1pt]
    \end{tabular}}}
    \hfill
    \subfloat{
    \scalebox{0.84}{
    \begin{tabular}{c|c c|c}
         \toprule[1pt]
         \multicolumn{4}{c}{UperNet}\\
         \midrule[0.5pt]
         Backbone & Params(M) & FLOPs(G) & mIoU(\%)\\
         \midrule[0.5pt]
         DAT-T~\cite{dat}& 60 & 957 & 45.5 \\
         NAT-T~\cite{NAT}& 58 & 934 & 47.1 \\
         InternImage-T~\cite{internimage}& 59 & 944 & 47.9 \\
         MPViT-S~\cite{mpvit}& 52 & 943 & 48.3 \\
         SMT-S~\cite{SMT}& 50 & 935 & 49.2 \\
         \rowcolor{gray!30}SSViT-S& 56 & 941 & \textbf{50.1} \\
         \midrule[0.5pt]
         DAT-S~\cite{dat}& 81 & 1079 & 48.3\\
         SMT-B~\cite{SMT}& 62 & 1004 & 49.6\\
         InterImage-S~\cite{internimage}& 80 & 1017 & 50.2\\
         MPViT-B~\cite{mpvit}& 105 & 1186 & 50.3 \\
         CSWin-S~\cite{cswin}& 65 & 1027 & 50.4\\
         \rowcolor{gray!30}SSViT-B& 86 & 1060 & \textbf{52.2}\\
         \midrule
         Swin-B~\cite{SwinTransformer} & 121 & 1188 & 48.1 \\
         GC ViT-B~\cite{globalvit} & 125 & 1348 & 49.2\\
         DAT-B~\cite{dat} & 121 & 1212 & 49.4 \\
         InternImage-B~\cite{internimage} & 128 & 1185 & 50.8\\
         CSWin-B~\cite{cswin} & 109 & 1222 & 51.1\\
         \rowcolor{gray!30}SSViT-L& 130 & 1256 & \textbf{53.3}\\
         \bottomrule[1pt]
    \end{tabular}}}
    \caption{Comparison with the state-of-the-art on ADE20K. }
    \vspace{-3mm}
    \label{tab:ade20k}
\end{table}

\paragraph{Results.} The results of semantic segmentation are detailed in Tab.~\ref{tab:ade20k}. All FLOPs are evaluated using an input resolution of $512\times2048$, with the exception of the SSViT-T group, which employs a $512\times 512$ resolution. Across all settings, SSViT delivers superior performance. Notably, within the Semantic FPN framework, our SSViT-S exceeds FAT-B3 by a significant \textbf{+0.7} mIoU margin. SSViT-L further outperforms CSWin-L by a remarkable \textbf{+1.6} mIoU. Within the UperNet framework, SSViT-S outstrips the recent SMT-S by \textbf{+0.9} mIoU. Both SSViT-B and SSViT-L also surpass their respective counterparts.


\subsection{Robustness Evaluation}

\paragraph{Settings.} In line with previous studies~\cite{RVT, FAN, MOAT}, we assess SSViT's robustness using ImageNet-V2~\cite{imagenetv2}, ImageNet-A~\cite{imagenet-a}, and ImageNet-R~\cite{imagenet-r}. The models used for this evaluation are pretrained on ImageNet-1k~\cite{imagenet}.

\paragraph{Results.} Tab.~\ref{tab:robustness} presents the robustness evaluation outcomes. On ImageNet-V2 (IN-V2), SSViT outperforms all competitors. For instance, SSViT-B exceeds BiFormer-B by \textbf{+1.7}, maintaining similar parameters and FLOPs. SSViT's advantages are further amplified on ImageNet-A (IN-A) and ImageNet-R (IN-R). Specifically, SSViT-L, pretrained solely on ImageNet-1k, achieves accuracies of \textbf{55.0} on IN-A and \textbf{59.2} on IN-R, markedly outpacing FAN-Hybrid-L (IN-A: \textbf{+13.2}, IN-R: \textbf{+6.0}). This underscores SSViT's robustness.



\begin{table}[h]
    \centering
    \setlength{\tabcolsep}{0.9mm}
    \subfloat{
    \scalebox{0.79}{
    \begin{tabular}{c|c c|c}
         \toprule[1pt]
         Backbone & Params(M) & FLOPs(G) & IN-V2(\%)\\
         \midrule[0.5pt]
         tiny-MOAT-2~\cite{MOAT} & 10 & 2.3 & 70.1 \\
         BiFormer-T~\cite{biformer} & 13 & 2.2 & 70.7 \\
         SMT-T~\cite{SMT} & 12 & 2.4 & 71.0 \\
         \rowcolor{gray!30}SSViT-T& 15 & 2.4 & \textbf{72.3}\\
         \midrule[0.5pt]
         XCiT-S12~\cite{xcit} & 26 & 4.8 & 72.5 \\
         SMT-S~\cite{SMT} & 21 & 4.7 & 73.3 \\
         BiFormer-S~\cite{biformer} & 26 & 4.5 & 73.6 \\
         \rowcolor{gray!30}SSViT-S& 27 & 4.4 & \textbf{74.1}\\
         \midrule[0.5pt]
         XCiT-S24~\cite{xcit} & 48 & 9.1 & 73.3 \\
         BiFormer-B~\cite{biformer} & 57 & 9.8 & 74.0 \\
         MOAT-1~\cite{MOAT} & 42 & 9.1 & 74.2 \\
         \rowcolor{gray!30}SSViT-B& 57 & 9.6 & \textbf{75.7} \\
         \midrule[0.5pt]
         DeiT-B~\cite{deit} & 86 & 17.5 & 71.5 \\
         MOAT-2~\cite{MOAT} & 73 & 17.2 & 74.3 \\
         \rowcolor{gray!30}SSViT-L& 100 & 18.2 & \textbf{76.1} \\
         \bottomrule[1pt]
    \end{tabular}}}
    \hfill
    \subfloat{
    \scalebox{0.79}{
    \begin{tabular}{c|c c|c c}
         \toprule[1pt]
         Backbone & Params(M) & FLOPs(G) & IN-A(\%) & IN-R(\%)\\
         \midrule[0.5pt]
         \rowcolor{gray!30}SSViT-T& 15 & 2.4 & \textbf{32.6} & \textbf{45.6}\\
         \midrule[0.5pt]
         Swin-T~\cite{SwinTransformer} & 29 & 4.5 & 21.6 & 41.3 \\
         ConvNeXt-T~\cite{convnext} & 29 & 4.5 & 24.2 & 47.2 \\
         RVT-S*~\cite{RVT} & 23 & 4.7 & 25.7 & 47.7 \\
         FAN-S-Hybrid~\cite{FAN} & 26 & 6.7 & 33.9 & 50.7 \\
         \rowcolor{gray!30}SSViT-S& 27 & 4.4 & \textbf{41.6} & \textbf{51.0}\\
         \midrule[0.5pt]
         ConvNeXt-S~\cite{convnext} & 50 & 8.7 & 31.2 & 49.5 \\
         LV-ViT-M~\cite{tokenlabel} & 56 & 16.0 & 35.2 & 47.2 \\
         FAN-B-Hybrid~\cite{FAN} & 50 & 11.3 & 39.6 & 52.9\\
         \rowcolor{gray!30}SSViT-B& 57 & 9.6 & \textbf{49.4} & \textbf{55.6}\\
         \midrule[0.5pt]
         Swin-B~\cite{SwinTransformer} & 88 & 15.4 & 35.8 & 46.6 \\
         MAE-ViT-B~\cite{MAE} & 86 & 17.5 & 35.9 & 48.3 \\
         RVT-B*~\cite{RVT} & 92 & 17.7 & 28.5 & 48.7 \\
         FAN-L-Hybrid~\cite{FAN} & 77 & 16.9 & 41.8 & 53.2 \\
         \rowcolor{gray!30}SSViT-L& 100 & 18.2 & \textbf{55.0} & \textbf{59.2}\\
         \bottomrule[1pt]
    \end{tabular}}}
    \caption{Evaluation of the model's robustness.}
    \vspace{-3mm}
    \label{tab:robustness}
\end{table}

\subsection{Ablation Study}

\begin{table*}[t]
    \centering
    \setlength{\tabcolsep}{1mm}
    \scalebox{1.00}{
    \begin{tabular}{c|c c|c|c c|c}
        \toprule[1pt]
         Model & Params(M) & FLOPs(G) & Top1-acc(\%) & $AP^b$ & $AP^m$ & mIoU(\%)\\
         \midrule
         DeiT-S~\cite{deit} & 22 & 4.6 & 79.8 & -- & -- & -- \\
         \rowcolor{gray!30}SS-DeiT-S & 22 & 4.3(\textcolor{red}{-0.3}) & 81.3(\textcolor{red}{+1.5}) & -- & -- & -- \\
         \midrule
         Swin-T~\cite{SwinTransformer} & 29 & 4.5 & 81.3 & 43.7 & 39.8 & 44.5 \\
         \rowcolor{gray!30}SS-Swin-T & 29 & 4.8 & 83.8(\textcolor{red}{+2.5}) & 47.8(\textcolor{red}{+4.1}) & 43.3(\textcolor{red}{+3.5}) & 49.3(\textcolor{red}{+4.8}) \\
         \midrule
         SSViT-T & 15 & 2.4 & \textbf{83.0} & \textbf{47.3} & \textbf{42.6} & \textbf{46.8} \\
         ${\rm S^3A}\xrightarrow{}$WSA~\cite{SwinTransformer} & 15 & 2.3 & 81.3 & 44.0 & 39.8 & 42.7\\
         ${\rm S^3A}\xrightarrow{}$CSWSA~\cite{cswin} & 15 & 2.3 & 81.6 & 44.6 & 40.3 & 43.7\\
         w/o LCE & 15 & 2.4 & 82.8 & 47.0 & 42.4 & 46.5 \\
         w/o CPE & 15 & 2.4 & 82.9 & 47.3 & 42.4 & 46.7 \\
         w/o Stem & 15 & 2.1 & 82.8 & 47.1 & 42.4 & 46.4 \\
         \bottomrule[1pt]
    \end{tabular}}
    \caption{Ablation study. }
    \vspace{-1mm}
    \label{tab:ablation_study}
\end{table*}

\paragraph{SSViT v.s. DeiT\& Swin.}
As illustrated in Tab.~\ref{tab:ablation_study}, we position SSViT in comparison with DeiT~\cite{deit} and Swin~\cite{SwinTransformer}. By exclusively substituting the Self-Attention/Window Self-Attention module in DeiT/Swin with ${\rm S^3A}$, we formulate SS-DeiT/SS-Swin-T. Remarkably, SS-DeiT-S, despite demanding fewer FLOPs, outperforms DeiT-S, registering a notable performance gain of \textbf{+1.5}, thereby underlining the potency of ${\rm S^3A}$. In contrast to Swin-T, SS-Swin-T exhibits substantial enhancements across a broad range of downstream tasks.

\paragraph{${\rm \mathbf{S^3A}}$ v.s. WSA.}
The Window Self-Attention (WSA), integral to the Swin Transformer~\cite{SwinTransformer}, is widely employed. In Tab.~\ref{tab:ablation_study}, we contrast ${\rm S^3A}$ with WSA, revealing ${\rm S^3A}$'s significant superiority in both classification and downstream tasks. Specifically, using SSViT-T as the benchmark, ${\rm S^3A}$ achieves a substantial \textbf{+1.7} boost in classification accuracy and a \textbf{+3.3} increase in box AP over WSA.


\paragraph{${\rm \mathbf{S^3A}}$ v.s. CSWSA.}
CSWSA, an enhancement of WSA introduced in CSwin-Transformer~\cite{cswin}, outperforms its predecessor. Yet, our ${\rm S^3A}$ still surpasses CSWSA in key performance metrics. Notably, ${\rm S^3A}$ attains a classification accuracy \textbf{+1.4} points higher than CSWSA. Within the Semantic FPN segmentation framework, ${\rm S^3A}$ exceeds CSWSA by a substantial \textbf{+3.1} mIoU.


\paragraph{LCE.} LCE, a straightforward depth-wise convolutional component, is employed to amplify the model's capacity to capture local features. We perform ablation studies on LCE, revealing its contribution to the model's performance enhancement. The results, presented in Tab.~\ref{tab:ablation_study}, indicate that LCE increases the model's classification accuracy by \textbf{+0.2}, box AP by \textbf{+0.3}, and mask AP by \textbf{+0.2}.


\paragraph{CPE.} CPE~\cite{CPVT} is a versatile, plug-and-play positional encoding strategy, frequently employed to impart positional information to the model. Comprising only a 3x3 depth-wise convolution in a residual block, CPE provides modest performance improvements as illustrated in Tab.~\ref{tab:ablation_study}, with an approximate increase of \textbf{+0.1} in classification accuracy.

\paragraph{Conv Stem.} The Conv Stem, deployed in the initial stages of the model, aids in extracting refined local features. Tab.~\ref{tab:ablation_study} suggests that the Conv Stem somewhat bolsters the model's performance in both classification and downstream tasks, specifically enhancing classification accuracy by \textbf{+0.2} and the mean Intersection over Union (mIoU) by \textbf{+0.4}.


\section{Conclusion}

Motivated by the human eye's efficient sparse scanning mechanism for visual information processing, we propose the Sparse Scan Self-Attention mechanism (${\rm S^3A}$). This mechanism emulates the human eye's procedural operation: initially selecting anchors of interest, subsequently extracting local information around these anchors, and ultimately aggregating this information. Harnessing the power of ${\rm S^3A}$, we develop the Sparse Scan Vision Transformer (SSViT), a robust vision backbone designed for a variety of vision tasks. We assess SSViT across a range of common visual tasks such as image classification, object detection, instance segmentation, and semantic segmentation, where it consistently showcases impressive performance. Notably, SSViT also exhibits remarkable robustness towards out-of-distribution (OOD) data.

\clearpage

{\small
\bibliographystyle{ieee_fullname}
\bibliography{egbib}
}
\newpage
\appendix
\section*{Supplementary Material}
In this supplementary material, we first provide the architecture details of our SSViT model, following by the experimental settings on different vision downstream tasks. And then we briefly discuss about the limitations and broader impacts of our study.

\begin{table}[ht]
    \centering
    \setlength{\tabcolsep}{0.8mm}
    \begin{tabular}{c|c c c c|c c}
         \toprule[1pt]
         Model & Blocks & Channels & Heads & Ratios & Params(M) & FLOPs(G)\\
         \midrule[0.5pt]
         SSViT-T & [2, 2, 9, 2] & [64, 128, 256, 512] & [2, 4, 8, 16] & 3 & 15 & 2.4 \\
         SSViT-S & [3, 5, 18, 4] & [64, 128, 256, 512] & [2, 4, 8, 16] & 3 & 27 & 4.4 \\
         SSViT-B & [4, 9, 25, 9] & [80, 160, 320, 512] & [5, 5, 10, 16] & 3 & 57 & 9.6 \\
         SSViT-L & [4, 9, 25, 9] & [112, 224, 448, 640] & [7, 7, 14, 20] & 3 & 100 & 18.2 \\
         \midrule[0.5pt]
         SS-Swin-T & [2, 2, 6, 2] & [96, 192, 384, 768] & [3, 6, 12, 24] & 4 & 29 & 4.8 \\
         SS-Swin-S & [2, 2, 18, 2] & [96, 192, 384, 768] & [3, 6, 12, 24] & 4 & 50 & 8.9 \\
         SS-Swin-B & [2, 2, 18, 2] & [128, 256, 512, 1024] & [4, 8, 16, 32] & 4 & 88 & 15.7 \\
         \midrule[0.5pt]
         SS-FasterViT-0 & [2, 3, 6, 5] & [64, 128, 256, 512] & [--, --, 8, 16] & 4 & 25 & 3.4 \\
         SS-FasterViT-1 & [1, 3, 8, 5] & [80, 160, 320, 640] & [--, --, 8, 16] & 4 & 41 & 5.4 \\
         SS-FasterViT-2 & [3, 3, 8, 5] & [96, 192, 384, 768] & [--, --, 12, 24] & 4 & 58 & 8.8 \\
         \bottomrule[1pt]
    \end{tabular}
    \caption{Detailed Architectures of our models.}
    \label{tab:arch}
\end{table}

\section{Architecture Details}

The architecture details are illustrated in Table~\ref{tab:arch}.
In SSViT, for the convolution stem, we adopt four $3\times 3$ convolutions to embed the input image into tokens. batch normalization and GELU are used after each convolution.
$3\times 3$ convolutions with stride 2 are used between stages to reduce the feature resolution.
$3\times 3$  depth-wise convolutions are adopted in CPE. While $5\times 5$ depth-wise convolutions are adopted for LCE. 

For SS-Swin, we strictly adhere to the design principles of the Swin-Transformer~\cite{SwinTransformer} without using additional structures such as CPE or Conv Stem.

For SS-FasterViT, we adhere to the design principles of FasterViT~\cite{fastervit}. We use convolutional structures in the first two stages of the model and apply ${\rm S^3A}$ in the latter two stages.

\section{Experimental Settings}

\paragraph{ImageNet Image Classification.}
We adopt the training strategy proposed in DeiT~\cite{deit}, but with the only supervision is classification loss. Specifically, our models are trained from scratch for 300 epochs with the input resolution of $224\times 224$. The AdamW is used with a cosine decay learning rate scheduler and 5 epochs of linear warm-up. The initial learning rate,  weight decay, and  batch-size are set to  0.001, 0.05, and 1024, respectively. We apply the same data augmentation and regularization used  in  DeiT~\cite{deit} (RandAugment \cite{randomaugment} (randm9-mstd0.5-inc1) , Mixup \cite{mixup} (prob = 0.8), CutMix \cite{cutmix} (prob = 1.0), Random Erasing (prob = 0.25), Exponential Moving Average (EMA) \cite{EMA}).
The  maximum rates of increasing stochastic depth \cite{droppath} are set to 0.1/0.15/0.4/0.5 for SSViT-T/S/B/L.

\paragraph{COCO Object Detection and Instance Segmentation.}
We apply RetinaNet~\cite{retinanet}, Mask-RCNN~\cite{maskrcnn}, and Cascaded Mask R-CNN~\cite{cai18cascadercnn} as the detection frameworks based on the MMDetection \cite{mmdetection}.
The models are trained unde ``1 $\times$" (12 training epochs) and ``3 $\times$ +MS" (36 training epochs with multi-scale training) settings. For the ``1 $\times$" setting, images are resized to the shorter side of 800 pixels while the longer side is within 1333 pixels. For the ``3 $\times$ +MS", multi-scale training strategy is applied to randomly resize the shorter side between 480 to 800 pixels.
We use the AdamW with the initial learning rate of 1e-4. For RetinaNet, we set the weight decay to 1e-4. While for Mask-RCNN and Cascaded Mask R-CNN, we set it to 5e-2.

\paragraph{ADE20K Semantic Segmentation.}Based on MMSegmentation~\cite{mmsegmentation}, we implement UperNet~\cite{upernet} and SemanticFPN~\cite{semanticfpn} to validate the SSViT. For UperNet, we follow the previous setting of Swin-Transformer~\cite{SwinTransformer} and train the model for 160k iterations with the input size of $512\times 512$. For SemanticFPN, we also use the input resolution of $512\times512$ but train the models for 80k iterations.

\paragraph{Robustness Evaluation.}Follow the previous works~\cite{FAN, RVT}, we evaluate the SSViT's robustness on the ImageNet-A~\cite{imagenet-a} and ImageNet-R~\cite{imagenet-r}. We also validated the model on ImageNet-V2~\cite{imagenetv2} to check for overfitting. All models are pretrained on ImageNet-1k.

\section{Limitations and Future Work.}
\label{app:limitation}
While the Sparse Scan Self-Attention mechanism (\(\rm{S}^3\rm{A}\)) and the Sparse Scan Vision Transformer (SSViT) have demonstrated significant computational efficiency and strong performance across various tasks, there are still several limitations that need to be addressed. One notable limitation is the computational constraints that prevented us from experimenting with larger models and datasets such as ImageNet-21k. Exploring the potential of SSViT on such large-scale datasets could provide further insights into its scalability and robustness. In the future, we will strive to validate the performance of SSViT on large datasets and with larger models.



\section{Broader Impact Statement.}
\label{app:imapact}
The development of the Sparse Scan Self-Attention mechanism (\(\rm{S}^3\rm{A}\)) and the Sparse Scan Vision Transformer (SSViT) has the potential to impact the field of computer vision by offering a more efficient alternative to traditional Transformers. By mimicking the human eye's sparse scanning mechanism, SSViT reduces computational load and improves the efficiency of vision models, which could lead to broader applications in resource-constrained environments. 



The proposed SSViT is a general vision backbone that can be applied on different vision tasks, e.g., image classification, object detection instance segmentation, and semantic segmentation. It has no direct negative social impact. Possible malicious uses of SSViT as a general-purpose backbone are beyond the scope of our study to discuss.

\section{Code}
\label{app:code}
We provide the code of our Sparse Scan Self-Attention.
\begin{lstlisting}

import torch.nn as nn
import torch
from einops import rearrange
from natten.functional import natten2dqkrpb, natten2dav

class S3A(nn.Module):

    def __init__(self, embed_dim, num_heads, window_size, anchor_size, stride):
        super().__init__()
        self.embed_dim = embed_dim
        self.num_heads = num_heads
        self.window_size = window_size
        self.anchor_size = anchor_size
        self.stride = stride
        self.head_dim = embed_dim // num_heads
        self.scaling = self.head_dim ** -0.5
        self.qkv = nn.Conv2d(embed_dim, embed_dim*3, 1, bias=True)
        self.out_proj = nn.Conv2d(embed_dim, embed_dim, 1, bias=True)

    def forward(self, x: torch.Tensor):
        '''
        x: (b c h w)
        '''
        bsz, _, h, w = x.size()
        qkv = self.qkv(x) # (b 3*c h w)

        q, k, v = rearrange(qkv, 'b (m n d) h w -> m b n h w d', m=3, n=self.num_heads)

        k = k * self.scaling

        window_size = self.window_size
        anchor_size = self.anchor_size
    
        attn = natten2dqkrpb(q, k, None, window_size, 1)
        attn = attn.softmax(dim=-1)
        v = natten2dav(attn, v, window_size, 1)

        stride = self.stride

        attn = natten2dqkrpb(q, k, None, anchor_size, stride)
        attn = attn.softmax(dim=-1)
        v = natten2dav(attn, v, anchor_size, stride)

        res = rearrange(v, 'b n h w d -> b (n d) h w')
        return self.out_proj(res)


\end{lstlisting}
\end{document}